\documentclass{article}

\usepackage[preprint]{neurips_2026}

\usepackage[utf8]{inputenc}
\usepackage[T1]{fontenc}
\usepackage{amsmath,amssymb,bm}
\usepackage{amsfonts}
\usepackage{graphicx}
\usepackage{booktabs}
\usepackage{array}
\usepackage{tabularx}
\usepackage{url}
\usepackage{hyperref}
\usepackage{microtype}
\usepackage{xcolor}

\graphicspath{{figures/}}

\newcommand{\method}{RGraph-Eval}
\newcommand{\sgscore}{\text{\method}}

\DeclareMathOperator{\median}{median}

\title{RGraph-Eval: Task-Grounded Scene Graphs
for Evaluating Robot Video Generation}

\author{
\textbf{Aleksandr Lazarev}$^{1,2}$ \quad
\textbf{Sofia Zimina}$^{1}$ \quad
\textbf{Roman Kuznetsov}$^{1}$ \\
\textbf{Tatiana Zemskova}$^{1,2}$ \quad
\textbf{Iaroslav Koshelev}$^{3}$ \quad
\textbf{Dmitry Yudin}$^{1,2}$ \\
$^{1}$MIRAI, Moscow, Russia \\
$^{2}$AXXX, Moscow, Russia \\
$^{3}$AI Foundation and Algorithm Lab, Moscow, Russia
}
\begin{document}

\maketitle

\begin{abstract}
Generated robot videos can look convincing while depicting the wrong interaction: a different object is moved, a grasp disappears, or a placement ends outside its destination. We introduce \method, a reference-based framework that makes these discrepancies explicit through task-grounded temporal scene graphs. Unlike approaches that construct separate evidence representations for different evaluation dimensions, we organize task entities, interactions, and motion within a unified, interpretable scene graph representation. A foundation-model pipeline identifies manipulation roles, grounds persistent object masks, infers and verifies state transitions and action intervals, and reconstructs manipulated-object motion in three dimensions. Applying this representation to a generated video and a reference demonstration supports an interpretable \method{} score with visible-object, relation-change, and 3D path-shape components. We also introduce ShareRobot-SG, a collection of 1,000 manipulation scenes from ten ShareRobot sources annotated with the multimodal representation. Our evaluation separates graph-generation quality from agreement with human video judgments, and includes comparisons with adapted SVG2 and SG-Ego, a benchmark of three video generators. The framework connects robot-video assessment to the observable entities and interactions underlying a task, while making the graph annotations reusable beyond evaluation. Code will be publicly available at \url{https://github.com/SashaDance/rgraph-eval}.
\end{abstract}

\section{Introduction}
\label{sec:introduction}
Video generation depicts possible robot behavior from an initial observation and a language goal, supporting behavior inspection, visual demonstrations, and data generation for robot learning~\citep{mei2026videogenerationrobotics}. Generated videos can serve as an additional source of training data, complementing demonstrations collected through physical interaction. Recent work explores scalable robot data synthesis with compositional world models~\citep{ye2026robodream} and learning robotic manipulation by imitating generated videos without physical demonstrations~\citep{patel2026rigvid}. However, using generated videos as a data source requires assessing their quality: visually realistic sequences may still depict incorrect object motion, inconsistent interactions, or failed task outcomes. Evaluation should therefore assess not only visual appearance but also whether the generated behavior preserves task-relevant entities, interactions, and final states.

Consider the instruction ``stack the white plate onto the red plate.'' A plausible video can move the wrong plate, omit the grasp, or leave the white plate beside the red one. Such errors may occupy a small region while the background remains stable. Conversely, different speeds or intermediate motions can preserve the intended interaction. Evaluation must distinguish appearance, object identity, temporal interaction, and the achieved state.

\begin{figure}[t]
\centering
\includegraphics[width=\columnwidth]{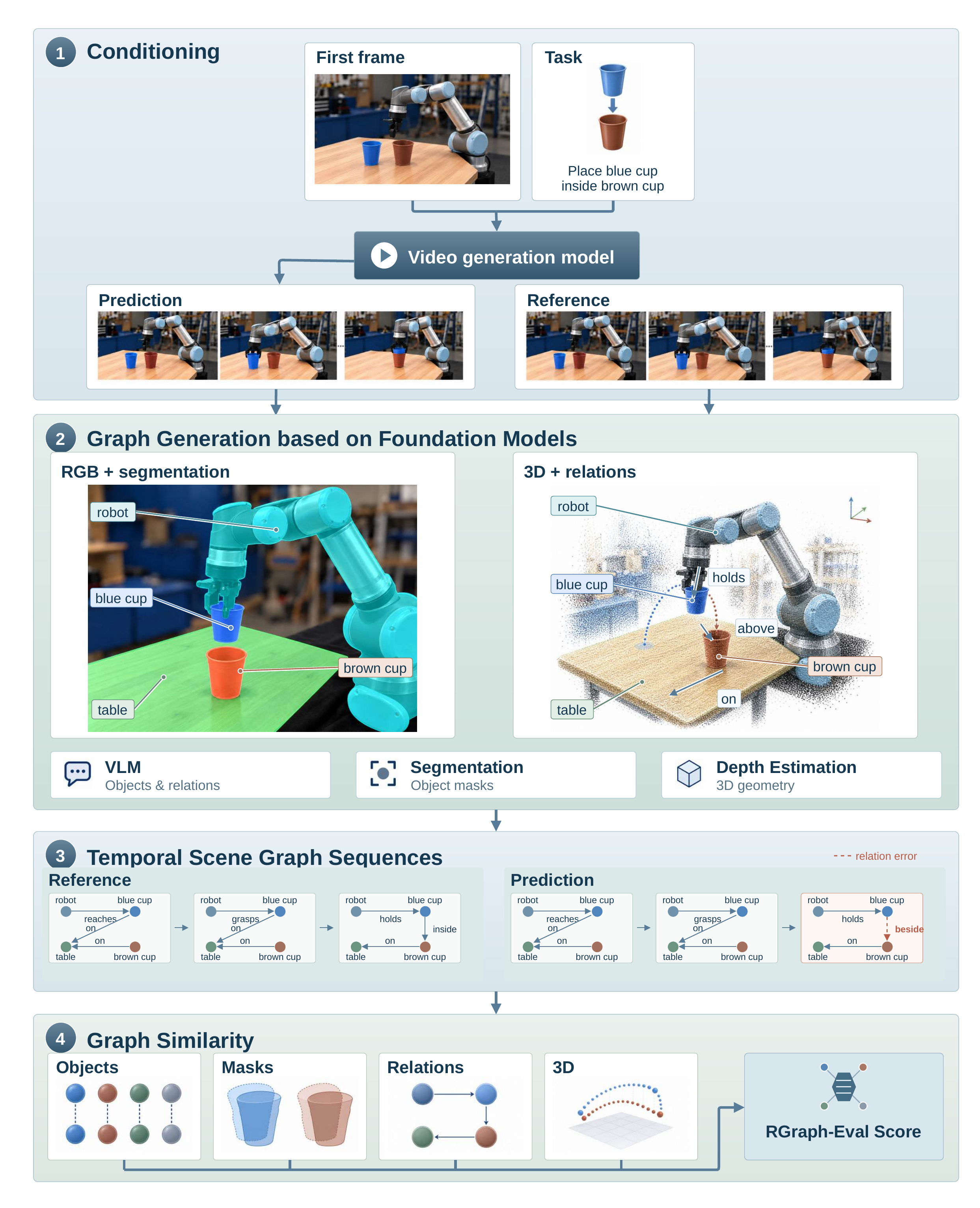}
\caption{\method\ overview. Foundation models transform reference and predicted videos into graph sequences. These scene graphs provide a structured summary of task-relevant entities and interactions and serve as the basis for evaluation. Visible-object, relation-change, and 3D path-shape comparisons contribute to the \method{} score.}
\label{fig:overview}
\end{figure}

Existing evaluators address complementary parts of this problem. General video benchmarks assess visual and temporal quality~\citep{vbench}. Robotics and world-model benchmarks emphasize instruction following and physical consistency~\citep{rbench,worldmodelbench}. Reference-based methods compare a prediction with an observed execution~\citep{ref4d}. Learned reward models additionally approximate human preferences or task progress~\citep{videoreward,roboreward,videoscore2}. These approaches provide useful assessments, but their scores need not identify which visible object or interaction accounts for a disagreement. A structural representation can supply that missing connection between an assessment and the evidence in the video.

Task-grounded temporal scene graphs connect assessment to this evidence. We call these graphs task-grounded because their entities and relations are selected with respect to the manipulation goal: they represent the robot, manipulated object, relevant supports and target, and their task-relevant interactions rather than exhaustively describing the entire scene. Nodes identify task entities, masks ground them in image regions, relations describe changing interactions, and trajectories capture motion. 

Graph-based assessment depends on extraction quality: incorrect object assignments propagate through tracking, relations, and geometry, while language models may infer goal-consistent rather than observed outcomes. We address these risks through visual grounding, persistent tracking, and independent verification, and evaluate graph quality separately from human alignment.

Figure~\ref{fig:overview} summarizes the framework. We make three contributions. First, we introduce a foundation-model pipeline for annotating manipulation videos with grounded temporal graphs, including masks, states, actions, and estimated 3D motion. Second, we present ShareRobot-SG, a 1,000-scene collection spanning ten manipulation sources for studying the quality and breadth of this representation. Third, we introduce reference-based \method{} assessment with an explicit component decomposition, evaluated through human alignment, generator comparison, and component analysis.

\section{Related Work}
\label{sec:related}
\subsection{Evaluating Generated Videos}
PSNR measures pixel-wise reconstruction error, SSIM compares local structure~\citep{ssim}, and LPIPS measures learned perceptual distance~\citep{lpips}. Frame-wise appearance similarity does not explicitly assess manipulated-object identity, interaction, or task completion~\citep{rbench,worldmodelbench}.

VBench evaluates appearance, consistency, and motion~\citep{vbench}. RBench emphasizes task adherence, physical plausibility, and subject stability~\citep{rbench}, while WorldModelBench assesses instruction following and world-model violations~\citep{worldmodelbench}. For manipulation, neither background stability nor a plausible final image establishes coherent robot--object interaction.

Human-aligned evaluators include TIQA, which learns text-quality prediction for generated images from human ratings~\citep{tiqa}, and VideoReward, which learns video preference dimensions~\citep{videoreward}. VideoScore2 assesses visual quality, text alignment, and physical consistency~\citep{videoscore2}, while RoboReward estimates robotic task progress~\citep{roboreward}. Ref4D-VideoBench compares reference and generated videos across semantics, events, and other dimensions~\citep{ref4d}. We instead make grounded objects, masks, relation changes, and geometry the comparison interface.

\subsection{Scene Graph Generation from Video}
Visual Genome and Action Genome establish object--relation representations for image and video understanding~\citep{visualgenome,actiongenome}. Recent annotation pipelines use foundation models to extend graph generation beyond fixed object categories. SVG2 extracts large-scale spatiotemporal scene graphs from video~\citep{svg2}. SG-Ego supports scene-based reasoning about human activities~\citep{sgego}. Their broad object vocabularies make them relevant comparisons for manipulation-focused extraction.

Our pipeline combines open-vocabulary object descriptions with task roles and an explicit state--action schema. We adapt SVG2 and SG-Ego to the same manipulation setting, although these methods naturally produce broad scene-description relations, including geometric relations. Such predicates are useful for general scene understanding, but they are not sufficient to determine whether a manipulation was executed correctly. Moreover, when ground-truth-quality depth and camera geometry are available, they can be computed directly from reconstructed 3D object positions.

\subsection{Scene Graphs in Robotics and Annotation}
Scene graphs already support robotic perception, planning, and control. VeriGraph uses graph structure to verify robot plans~\citep{verigraph}, and Compose by Focus uses task-relevant scene graphs for skill composition~\citep{composefocus}. SG-RAPL uses a 3D scene graph to ground language-driven, hierarchical action planning for humanoid robots in warehouse environments~\citep{sgrapl}. These works establish that structured object relations have value beyond image description.

Despite these applications, datasets pairing robotic manipulation observations with explicit scene-graph annotations remain relatively scarce. SG-Bot provides a synthetic tabletop dataset with graph, depth, and segmentation annotations~\citep{sgbot}. EmbodimentSemantic pairs real-robot observations with spatial scene graphs and provides a simulator-grounded manipulation benchmark~\citep{embodimentsemantic}.

ShareRobot-SG augments ShareRobot manipulation videos~\citep{sharerobot} with object identities, masks, changing relations, action timings, and estimated 3D motion. 
\section{Method}
\label{sec:method}
Each evaluation uses a task goal $g$, a reference demonstration $V^r$, and a predicted video $V^p$ generated from an initial image and the goal. The reference shows one valid execution, not all possible solutions. The same extractor $\Phi$ processes each video independently:
\begin{equation}
\mathcal{G}^r=\Phi(V^r,g),\qquad \mathcal{G}^p=\Phi(V^p,g).
\label{eq:extract}
\end{equation}
\begin{figure}[t]
\centering
\includegraphics[width=0.99\textwidth]{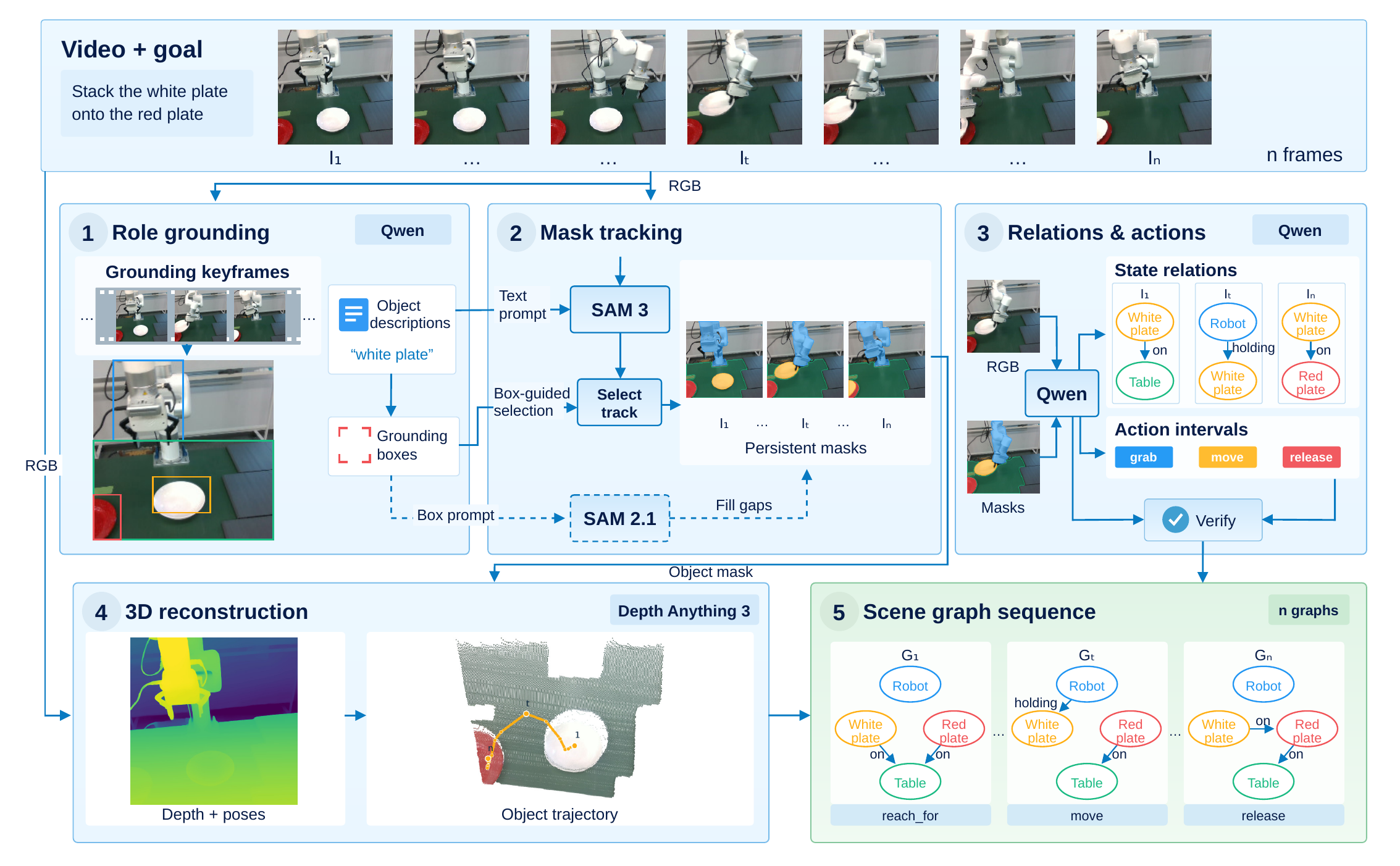}
\caption{Scene graph generation for a UTokyo xArm episode. Qwen3.8-27B~\citep{qwen} identifies entities and grounding keyframes. SAM 3 generates text-prompted tracks, boxes select the intended objects, and SAM 2.1 fills gaps. Qwen3.8-27B infers and verifies states and actions. Depth and camera estimates provide object motion, resulting in a multimodal temporal scene graph sequence.}
\label{fig:pipeline}
\end{figure}

The pipeline takes the goal and ordered RGB frames as input and follows the five stages in Fig.~\ref{fig:pipeline}. It selects grounding keyframes internally and requires no external masks, hand-labelled boxes, or demonstration planning steps. Masks link graph entities to image regions, and frame indices link prepared frames to the original video.

\subsection{Role Grounding}

We ground four core task roles: robot, manipulated object, initial support, and target, with an optional whole-parent entity for part--whole interactions. Qwen3.8-27B~\citep{qwen} uses the goal and ordered frames to describe each entity and predict grounding boxes at selected keyframes. These descriptions provide segmentation prompts, while the boxes disambiguate visually similar objects. A verification pass checks the proposed entities against the frames, and visual evidence takes precedence when it conflicts with the goal.

\subsection{Mask Tracking}
The grounded descriptions and boxes define which objects to track. SAM~3~\citep{sam3} uses these descriptions as text prompts to generate candidate mask tracks from the RGB sequence. Each prompt may produce several candidate tracks when objects look similar. We select the intended tracks using overlap between their masks and the grounding boxes from Qwen3.8-27B. The boxes are used after segmentation to select tracks, not as prompts to SAM~3.

SAM~2.1~\citep{sam2} takes RGB frames and grounding boxes as prompts, then propagates masks within temporal chunks. It fills only frames where the selected SAM~3 track has no mask and keeps existing SAM~3 masks. The dashed branch in Fig.~\ref{fig:pipeline} shows this gap-filling step.

For each entity, we store masks, image-space boxes, centroids, areas, and visibility to link its graph node to the video.

\subsection{Relations and Actions}
Once objects are identified and tracked, we add relations and actions to describe their interactions. Relations represent interaction states, while actions describe events over time. For example, \texttt{holding} is a robot--object relation, while \texttt{grab} and \texttt{release} mark changes in that relation. We keep states and actions separate.

From the ordered frames, Qwen3.8-27B predicts initial relations, state changes, action intervals, and final relations. If $\Delta_t^-$ and $\Delta_t^+$ are the predicates removed and added at observation $t$, we update the state as
\begin{equation}
\mathcal{R}_t=(\mathcal{R}_{t-1}\setminus\Delta_t^-)\cup\Delta_t^+.
\label{eq:state}
\end{equation}
An action with inclusive interval $[s,e]$ belongs to $\mathcal{A}_t$ when $s\leq t\leq e$. 

A second pass with Qwen3.8-27B checks the states, transitions, and actions against RGB frames and entity-mask overlays. It verifies object identity, contact and lift timing, and the final placement. The verifier can remove unsupported facts instead of assuming that the goal was achieved.

State relations include support, containment, holding, touching, part--whole and attachment relations, and open/closed states. Actions include reaching, grasping, lifting, moving, placing, releasing, pushing, pouring, opening, closing, insertion, and stacking. We use this fixed interaction schema with open-vocabulary object descriptions. Relations are inferred from visual evidence.

\subsection{3D Reconstruction}
Beyond relations and actions, we add 3D information that, combined with the tracked masks, yields an estimated object trajectory. We focus this reconstruction on the manipulated object. Depth Anything~3~\citep{da3} estimates depth $D_t$, camera intrinsics $K_t$, and a camera-to-world transform $T_t$. The manipulated-object mask selects pixels with valid positive depth, $\Omega_t$. For homogeneous pixel coordinates $\mathbf{u}=(u,v,1)^\top$, we back-project these pixels to estimate the object's position:
\begin{align}
\mathbf{c}_t &= \median_{(u,v)\in\Omega_t}\!\left[D_t(u,v)K_t^{-1}\mathbf{u}\right],\\
\begin{bmatrix}\mathbf{x}_t\\1\end{bmatrix}
&=T_t\begin{bmatrix}\mathbf{c}_t\\1\end{bmatrix}.
\label{eq:trajectory}
\end{align}
We take the component-wise median in camera coordinates, then transform it to the reconstructed world frame. This tracks the visible object surface, not its full pose or center of mass. Empty masks or invalid depth leave the position unavailable.

We use estimated 3D motion to penalize ambiguous or inconsistent object movement relative to the reference, rather than to recover an exact physical trajectory. It also helps inspect relations inferred from ambiguous images. Because separate reconstructions can differ in scale and coordinate frame, 3D comparison requires compatible coordinates and normalization.

\subsection{Scene Graph Sequence}
We combine the tracked entities and masks, verified relations and action intervals, and estimated 3D positions into one graph per prepared frame.
For $n$ prepared video frames, the graph sequence is $\mathcal{G}=(G_1,\ldots,G_n)$, with
\begin{equation}
G_t=(\mathcal{V}_t,\mathcal{M}_t,\mathcal{R}_t,\mathcal{A}_t,\mathbf{x}_t).
\label{eq:graph}
\end{equation}
These terms represent task entities, masks, state relations, active actions, and the manipulated object's estimated 3D position. Entity IDs remain fixed within each video.

The same graph-generation pipeline processes reference and generated videos independently. The goal provides task context, while states and actions are inferred from RGB frames and mask overlays; visual evidence takes precedence when it conflicts with the goal. This prevents the graph from simply reflecting the intended task and allows \method{} to capture execution errors in generated videos. The same principle also handles occasional mismatches between ShareRobot planning goals and recorded executions.

\subsection{Reference-Based \method{} Score}
\label{sec:score}
We compare generated ($p$) and reference ($r$) graphs at up to 30 uniformly spaced relative-progress samples, using visible-object agreement, relation changes, and manipulated-object 3D path shape.

\paragraph{Object correspondence}
We establish a fixed one-to-one correspondence $\pi$ between predicted
and reference objects using semantic identity and visual evidence in
the observation prefix. Role labels alone do not establish a match.
The same correspondence is used for all three components.

\paragraph{Visible-object agreement}
Let $B_i^v$ contain the non-robot objects marked visible at paired
frame $i$ in video $v$. We compute
\begin{equation}
O=\frac{2\sum_i |B_i^p\cap_{\pi} B_i^r|}
{\sum_i (|B_i^p|+|B_i^r|)},
\label{eq:score_objects}
\end{equation}
where $\cap_{\pi}$ counts identity-matched objects visible in both
frames. Missing, extra, or incorrectly matched objects reduce
agreement. If both sets are empty throughout, $O=1$.
This term measures object presence, not mask overlap.

\paragraph{Relation-change agreement}
For $k\in\{\mathrm{state},\mathrm{action}\}$, let $E_k^v$
be the chronological sequence of relation additions and removals
involving the manipulated object. Events match when their change
types, predicates, and identity-matched endpoints agree. We define
\begin{equation}
R=\frac{1}{2}\sum_k
\frac{2\,\operatorname{LCS}_{\pi}(E_k^p,E_k^r)}
{|E_k^p|+|E_k^r|},
\label{eq:score_relations}
\end{equation}
where $\operatorname{LCS}_{\pi}$ counts the longest sequence of
matching events that preserves their order in both videos.

\paragraph{3D path-shape agreement}
We compare trajectories of the identity-matched manipulated object.
Of $N$ progress-aligned position pairs, let $m$ be valid in both
videos. If no object match exists or $m<2$, set $T=0$.
Otherwise, normalize positions by the object's first valid
positive median depth $d^v$:
\begin{equation}
\mathbf{z}_i^v=
\frac{\mathbf{x}_i^v-\mathbf{x}_1^v}{d^v},
\qquad
D_{ij}^v=\|\mathbf{z}_i^v-\mathbf{z}_j^v\|_2.
\label{eq:score_shape}
\end{equation}
With $s=\max(\max_i\|\mathbf{z}_i^r\|_2,10^{-6})$, define
\begin{equation}
T=\frac{m}{N}\exp\!\left(
-\frac{\sum_{i,j=1}^{m}|D_{ij}^p-D_{ij}^r|}{m^2s}
\right).
\label{eq:score_trajectory}
\end{equation}
Coverage $m/N$ penalizes missing observations. This term compares
motion shape, independent of absolute position and orientation;
it does not verify the final destination.

The final score is
\begin{equation}
S_{\sgscore}(V^p,V^r,g)=\frac{O+R+T}{3}.
\label{eq:score}
\end{equation}
All components lie in $[0,1]$; higher is better.

\section{Experiments}
\label{sec:experiments}
We evaluate graph-generation quality and tracking on ShareRobot-SG. Separately, we use the 30-scene generation subset to assess agreement with human video judgments, compare generators, and analyze the \method{} components.

\subsection{Datasets}
\label{sec:dataset}

Our experiments use two subsets derived from ShareRobot~\citep{sharerobot,robobrain}.
First, \textbf{ShareRobot-SG} contains 1,000 manipulation scenes selected from
ten ShareRobot source datasets. Of these, 900 scenes contain automatically
generated scene-graph annotations, while 100 scenes additionally have manual
ground-truth annotations used to evaluate graph-generation quality. Each scene
corresponds to an episode-level video paired with a task goal and is annotated
with task entities, persistent masks, temporal states and actions, depth and
camera estimates, and a manipulated-object trajectory.

Second, the video-generation experiments use a separate \textbf{30-scene
generation subset} drawn from five ShareRobot source datasets. These scenes are
used as common reference episodes for evaluating the compared video-generation
models. Table~\ref{tab:datasets} summarizes the composition of both subsets.

\begin{table}[t]
\centering
\caption{Dataset composition used in the experiments. ShareRobot-SG contains
1,000 scenes, including 100 manually annotated ground-truth scenes. The
video-generation evaluation uses a separate 30-scene subset.}
\label{tab:datasets}
\footnotesize
\setlength{\tabcolsep}{3pt}
\begin{tabularx}{\columnwidth}{@{}lXrr@{}}
\toprule
Dataset & Tasks & SG & Gen. \\
\midrule
UTokyo
    & Pick/place, stacking
    & 92 & 7 \\

DLR SARA Grid
    & Pick/place
    & 40 & -- \\

DLR SARA Pour
    & Pouring
    & 51 & -- \\

DLR EDAN
    & Picking
    & 63 & -- \\

UCSD Pick/Place
    & Pick/place
    & 151 & 5 \\

VIOLA
    & Relocation
    & 44 & -- \\

Bridge
    & Placement, relocation
    & 151 & 6 \\

PLEX
    & Placement, stacking, insertion, opening
    & 151 & -- \\

RoboSet
    & Pick/place, dragging, opening/closing
    & 151 & -- \\

ASU
    & Pick/place, pushing
    & 106 & -- \\

Berkeley Autolab
    & Pick/place
    & -- & 7 \\

UCSD Kitchen
    & Manipulation
    & -- & 5 \\

\midrule
\textbf{Total}
    & 
    & \textbf{1,000}
    & \textbf{30} \\
\bottomrule
\end{tabularx}
\end{table}

\begin{figure}[t]
\centering
\includegraphics[width=1.0\textwidth]{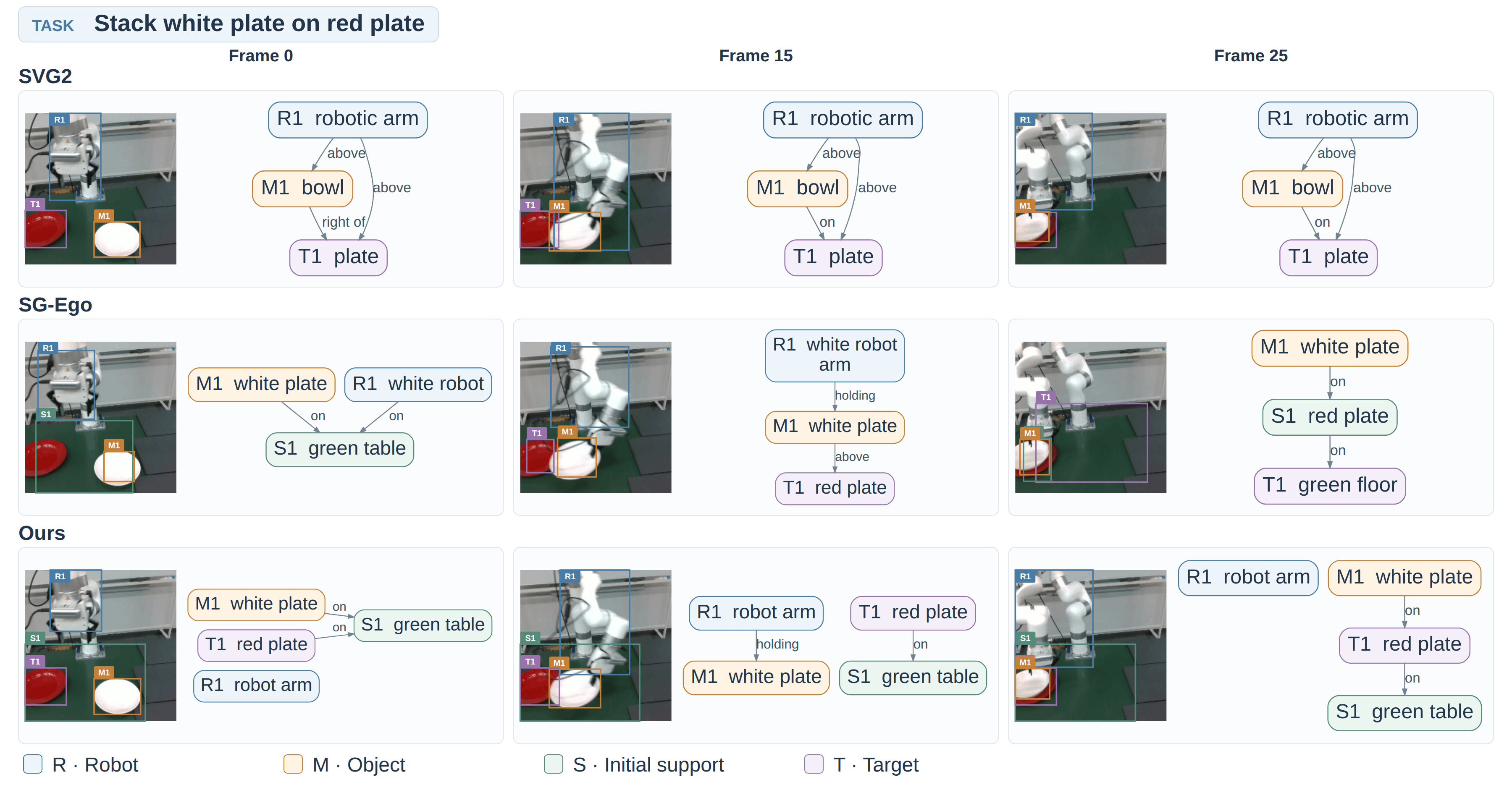}
\caption{Qualitative graph comparison on a UTokyo xArm plate-stacking episode. SVG2 and SG-Ego include broad geometric predicates alongside interaction relations, while ours primarily represents task roles and manipulation-relevant relations. In our graph, the robot holds the white plate in the intermediate frame, and the white plate is supported by the red plate in the final frame. Boxes ground entities and colors denote task roles.}
\label{fig:graph_comparison}
\end{figure}

\subsection{Comparison of Scene Graph Generation}
\label{sec:graph_comparison}
We compare our pipeline with adapted SVG2~\citep{svg2} and SG-Ego~\citep{sgego} on ShareRobot-SG, evaluating robot, manipulated-object, initial-support, and target roles only. The adapted SVG2 pipeline uses Qwen3.8-27B~\citep{qwen}. Table~\ref{tab:graph} reports object-role and relation precision, recall, and F1 (percent).

\paragraph{Evaluation protocol}
\label{sec:evaluation_protocol}
The 100 manually annotated videos in ShareRobot-SG comprise 80 primary videos sampled proportionally by dataset--task strata and 20 disagreement-selected challenge videos. Invalid goal--video pairs and judge-indeterminate cases are excluded. Only the probability-sampled primary subset calibrates VLM error; challenge cases do not contribute to this correction.

Object-role evaluation checks whether each predicted track follows the correct physical object; an incorrect required-role track incurs both FP and FN. We micro-average counts across videos. For a common relation scope across our method, SVG2, and SG-Ego, we evaluate only \texttt{holding}/\texttt{touching}, \texttt{on}/\texttt{supporting}, \texttt{inside}, and \texttt{attached\_to} edges, including their synonyms. All other relation types are excluded from the comparison for every method. Synonymous predicates are merged into canonical categories.

\paragraph{Prediction-powered estimates}
Gemma~4 (31B)~\citep{gemmateam2026gemma4technicalreport} serves as the evaluation judge and assesses up to ten distributed frames against candidate facts pooled from all methods and an independent proposer. Yes/uncertain/no map to $1/0.5/0$, yielding auxiliary TP/FP/FN counts; unshown frames receive no interpolated labels. We use prediction-powered inference~\citep{angelopoulos2023predictionpowered}: dataset-level VLM means are corrected by inverse-inclusion-weighted human-primary residuals, then combined by population weights. TP and actual-positive counts are corrected; predicted-positive counts are exact. Corrected moments are projected to feasible counts before computing P/R/F1.

\begin{table}[t]
\centering
\caption{Graph-generation quality (\%) on ShareRobot-SG and estimated latency per video (mean $\pm$ SD, seconds). Human micro-averages and dataset-stratified PPI estimates are distinct evaluations. P and R denote precision and recall. Bold marks the highest quality score within each evaluation group.}
\label{tab:graph}
\footnotesize
\setlength{\tabcolsep}{2pt}
\begin{tabular*}{\columnwidth}{@{\extracolsep{\fill}}lccccccc@{}}
\toprule
& \multicolumn{3}{c}{Objects} & \multicolumn{3}{c}{Relations} & \\
\cmidrule(lr){2-4}\cmidrule(lr){5-7}
Method & P & R & F1 & P & R & F1 & Latency \\
\midrule
\multicolumn{8}{@{}l}{\textit{Human evaluation}}\\
\addlinespace[2pt]
\textbf{Ours} & 90.77 & 88.59 & 89.67 & \textbf{94.04} & 73.74 & \textbf{82.66} & $94 \pm 24$\\
SG-Ego & \textbf{91.37} & \textbf{92.19} & \textbf{91.78} & 55.97 & \textbf{85.97} & 67.80 & $34 \pm 8$\\
SVG2 & 71.67 & 51.65 & 60.03 & 75.00 & 19.42 & 30.86 & $362 \pm 76$\\
\midrule
\multicolumn{8}{@{}l}{\textit{Dataset-stratified PPI}}\\
\addlinespace[2pt]
\textbf{Ours} & \textbf{93.19} & \textbf{90.55} & \textbf{91.85} & \textbf{82.61} & 64.30 & \textbf{72.31} & $94 \pm 24$\\
SG-Ego & 65.69 & 64.34 & 65.01 & 31.25 & \textbf{85.33} & 45.75 & $34 \pm 8$\\
SVG2 & 68.21 & 50.04 & 57.73 & 53.70 & 18.07 & 27.04 & $362 \pm 76$\\
\bottomrule
\end{tabular*}
\end{table}

Latencies sum processing for 30 sampled frames on A100 80\,GB GPUs, excluding model loading.

Overall, our method provides stronger manipulation-relation extraction while maintaining competitive object-role grounding. In addition, \method{} produces persistent object masks with consistent entity IDs across frames, linking graph entities to image regions over time and supporting the object trajectories used by the \method{} score. The tracking ablation in Table~\ref{tab:tracking_ablation} further shows the importance of this persistent grounding.

Figure~\ref{fig:graph_comparison} illustrates this difference in emphasis. SVG2 and SG-Ego describe several instantaneous geometric relationships between visible objects. These relations provide general spatial context, but they do not directly establish which object is manipulated, whether the robot maintains contact with it, or whether it reaches the intended target. Table~\ref{tab:graph} reports aggregate quality under the common manipulation-relation scope.

\subsection{Video Generation Evaluation}
\label{sec:human}

One assessor rated 90 videos generated from the 30-scene generation subset
(30 videos per generator) on task completion, object consistency, and visual
quality, each from 1 to 5. For each video, the overall human rating is the
arithmetic mean of its task-completion, object-consistency, and visual-quality
ratings, with equal weight for all three. Following a correlation-evaluation
protocol similar to RBench~\citep{rbench}, we average each automatic score and
each human-rating measure over the 30 videos per generator and compute Pearson's
$r$ across the three generator means. The Mean column in
Tables~\ref{tab:human} and~\ref{tab:component_analysis} reports correlation
with this overall human rating, not the average of the correlations for the
individual criteria. Generation is conditioned on planning
annotations from ShareRobot, because these annotations can be ambiguous and may not uniquely specify the demonstrated execution.

\begin{table}[t]
\centering
\caption{Pearson correlation across three generator means on the 30-scene generation subset (90 generated videos, one assessor). Rows are sorted by Mean. Bold marks the highest value in each column.}
\label{tab:human}
\footnotesize
\setlength{\tabcolsep}{2pt}
\begin{tabularx}{\columnwidth}{@{}Xcccc@{}}
\toprule
Evaluator & \shortstack{Task\\completion} & \shortstack{Object\\consistency} & \shortstack{Visual\\quality} & Mean\\
\midrule
\textbf{\method{} (ours)} & \textbf{0.9800} & 0.8633 & \textbf{0.9613} & \textbf{0.9722}\\
RBench: overall~\citep{rbench} & \textbf{0.9800} & 0.8098 & 0.9298 & 0.9447\\
Ref4D: event~\citep{ref4d} & 0.7028 & \textbf{0.9757} & 0.8912 & 0.8710\\
WorldModelBench~\citep{worldmodelbench} & 0.9078 & 0.5373 & 0.7293 & 0.7579\\
RoboReward~\citep{roboreward} & 0.8737 & 0.4725 & 0.6758 & 0.7067\\
VideoScore2: overall~\citep{videoscore2} & 0.5879 & 0.0577 & 0.3029 & 0.3434\\
VideoReward: overall~\citep{videoreward} & 0.5639 & 0.0284 & 0.2748 & 0.3156\\

\bottomrule
\end{tabularx}
\end{table}

Table~\ref{tab:human} compares author-supported aggregate scores, retaining
Ref4D's event score separately. \method{} and RBench tie for the highest
reported task-completion correlation, while \method{} shows the strongest
correlation with visual quality and the overall human rating. Ref4D's event
score correlates most strongly with object consistency. These results suggest
that the structured representation used by \method{} captures complementary
aspects of human judgment: relation changes directly reflect whether the
intended manipulation is carried out, while persistent object grounding and
3D motion provide additional evidence about object consistency.

\begin{figure}[t]
\centering
\includegraphics[width=1.0\textwidth]{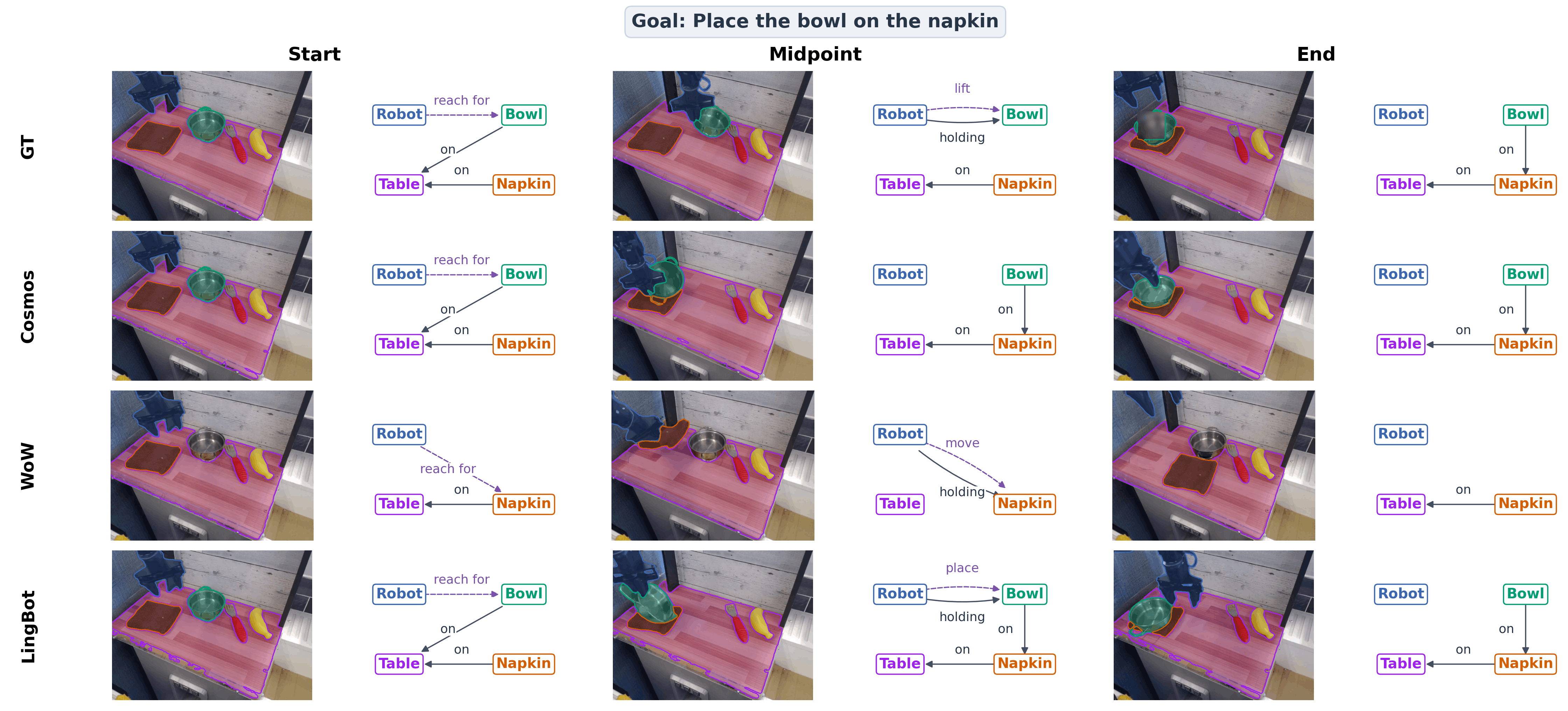}
\caption{Reference and generated Bridge bowl-placement videos from Cosmos, WoW, and LingBot at the start, midpoint, and end. Colored masks identify objects; adjacent graphs show relations (solid arrows) and actions (dashed arrows). The goal is to place the bowl on the napkin. Cosmos and LingBot place the bowl on the napkin, whereas WoW moves the napkin instead, leaving the bowl on the table.}
\label{fig:sushi_comparison}
\end{figure}

We further use the same 30-scene generation subset to compare the three video
generation models directly. Table~\ref{tab:generators} compares
Cosmos~3 Nano~\citep{cosmos_model}, LingBot-VA~\citep{lingbot_va}, and
WoW-1-Wan-14B-2M~\citep{wow_model}. LingBot-VA, the largest model among the
three, shows the strongest agreement in manipulation relations and 3D object
motion, suggesting better preservation of the task dynamics captured by
\method{}. Cosmos~3 Nano performs best on visible-object agreement, indicating
strong preservation of task-relevant entities, while WoW-1 also captures
object-motion structure well despite lower overall agreement. These results
highlight that the generators differ not only in overall quality, but in which
aspects of the manipulation sequence they preserve most faithfully.

Figure~\ref{fig:sushi_comparison} illustrates a task-level failure across the compared generators. The goal is to place the bowl on the napkin: Cosmos and LingBot achieve the intended final relation, whereas WoW moves the napkin instead and leaves the bowl on the table.

\begin{table}[t]
\centering
\caption{Generator means on the 30-scene generation subset: all components and the \method{} score lie in $[0,1]$. Rows are sorted by \method{}. Bold marks the highest value in each column.}
\label{tab:generators}
\small
\setlength{\tabcolsep}{3pt}
\begin{tabularx}{\columnwidth}{@{}Xcccc@{}}
\toprule
Generator & $O$ & $R$ & $T$ & \textbf{\method{}}\\
\midrule
LingBot-VA~\citep{lingbot_va} & 0.9928 & \textbf{0.9448} & \textbf{0.7988} & \textbf{0.9121}\\
Cosmos 3 Nano~\citep{cosmos_model} & \textbf{0.9983} & 0.9251 & 0.7608 & 0.8947\\
WoW-1~\citep{wow_model} & 0.9685 & 0.8798 & 0.7880 & 0.8788\\
\bottomrule
\end{tabularx}
\end{table}

\subsection{Ablation Studies}
\label{sec:ablations}

We conduct two ablations to examine the main design choices of \method{}:
the tracking pipeline used to construct persistent object representations and
the components used to compute the final evaluation score.

\paragraph{Tracking pipeline}
The tracking ablation isolates the effects of semantic prompting and temporal
mask repair. We compare four tracking variants on 100 valid annotated videos
from ShareRobot-SG, using the same ten frames per video and fixed Qwen
proposals. Removing SAM2.1 missing-frame repair substantially reduces tracking
recall, showing that temporal repair is important for maintaining object
visibility across the sequence. Box-prompted SAM2.1 provides strong precision,
but does not recover the recall of the full pipeline, while box-prompted SAM3
performs worse overall. Together, these results indicate that text-based
semantic prompting and SAM2.1 temporal repair provide complementary benefits
for persistent object grounding.

\begin{table}[t]
\centering
\caption{Role-object tracking ablation on 100 annotated ShareRobot-SG videos using the same ten frames per video and fixed Qwen proposals. P, R, and F1 are percentages; bold marks the highest value in each column. Repair denotes SAM2.1 missing-frame repair.}
\label{tab:tracking_ablation}
\small
\setlength{\tabcolsep}{3pt}
\begin{tabularx}{\columnwidth}{@{}Xccc@{}}
\toprule
Variant & P & R & F1 \\
\midrule
\textbf{Text-prompted SAM3 + repair (full)} & 89.65 & \textbf{89.51} & \textbf{89.58} \\
Text-prompted SAM3 & 89.41 & 72.86 & 80.29 \\
Box-prompted SAM3 & 82.25 & 76.70 & 79.38 \\
Box-prompted SAM2.1 & \textbf{90.54} & 84.02 & 87.16 \\
\bottomrule
\end{tabularx}
\end{table}

\paragraph{Score components}
The score ablation examines the contribution of visible-object agreement ($O$),
relation changes ($R$), and 3D path shape ($T$) using the same 90 generated
videos and human ratings from the 30-scene generation subset.
Relation changes provide the strongest standalone alignment with task
completion, indicating that changes in manipulation state are particularly
informative for judging whether the intended interaction is performed.
Visible objects and 3D motion provide complementary information, and combining
components generally improves alignment with human judgments. The full score
shows the strongest task-completion alignment, whereas the combination of
relations and 3D path shape aligns most strongly with the overall human rating.
This suggests that the components capture different aspects of generated
robot behavior.

\begin{table}[t]
\centering
\caption{Score ablations: Pearson $r$ across three generator means on the 30-scene generation subset. $O$: visible-object F1; $R$: relation changes; $T$: 3D path shape. Bold marks the highest value per column.}
\label{tab:component_analysis}
\small
\begin{tabularx}{\columnwidth}{@{}Xcc@{}}
\toprule
Variant & \shortstack{Task\\completion} & Mean\\
\midrule
Visible objects ($O$) & 0.7756 & 0.5726\\
Relation changes ($R$) & 0.9786 & 0.8845\\
3D path shape ($T$) & 0.2591 & 0.5141\\
\midrule
$(O+T)/2$ & 0.8889 & 0.9805\\
$(O+R)/2$ & 0.9362 & 0.8038\\
$(T+R)/2$ & 0.9523 & \textbf{0.9812}\\
\midrule
\textbf{Full} $\boldsymbol{(O+R+T)/3}$ & \textbf{0.9800} & 0.9722\\
\bottomrule
\end{tabularx}
\end{table}

\section{Conclusion and Future work}
\label{sec:conclusion}
We introduced \method, a reference-based evaluator that represents robot videos with task-grounded scene graphs combining object grounding, interaction changes, and 3D motion. We also introduced ShareRobot-SG, a 1,000-scene dataset with multimodal scene-graph annotations.

Across the video-generation benchmark, \method{} shows strong agreement with human judgments and captures complementary aspects of generated behavior through visible objects, relation changes, and object motion. The component analysis further indicates that these signals contribute differently to task-completion and overall human alignment.

Extraction errors and single-reference comparison remain limitations, and valid alternative executions may receive lower scores. The current representation also focuses on a single manipulated object and does not certify executable actions. Within this scope, \method{} provides an interpretable evaluation based on the entities and interactions underlying the task.

We plan to use object, relation, and motion discrepancies for iterative video refinement. Future work will also consider multi-object and bimanual representations, multi-reference comparisons, uncertainty-aware scoring, and evaluation on a larger benchmark covering a broader range of video-generation models, including closed-source systems.

\section*{Acknowledgments}
This work was partially carried out as part of the research project ``Improving Video-Based Visual-Spatial Question Answering Using 3D Scene Graphs'' under an agreement with Innopolis university.

\bibliographystyle{plainnat}
\bibliography{references}

\end{document}